\documentclass{article}
\usepackage{times}
\usepackage{graphicx} 
\usepackage{subfigure} 

\usepackage{natbib}

\usepackage{algorithm}
\usepackage{algorithmic}

\usepackage{url}

\usepackage[bottom]{footmisc}
\usepackage{verbatim} 
\usepackage[utf8]{inputenc}
\usepackage{amsfonts}
\usepackage{wrapfig}
\usepackage[titletoc]{appendix}
\usepackage{amsmath}
\usepackage{amssymb}
\usepackage{graphicx}
\usepackage{enumerate}
\usepackage{amsthm}
\usepackage{etoolbox}
\usepackage{picins}
\usepackage{lipsum}

\newtheorem*{rep@theorem}{\rep@title}
\newcommand{\newreptheorem}[2]{%
\newenvironment{rep#1}[1]{%
 \def\rep@title{#2 \ref{##1}}%
 \begin{rep@theorem}}%
 {\end{rep@theorem}}}
\makeatother

\newtheorem{thm}{Theorem}
\newreptheorem{thm}{Theorem}
\newtheorem{lem}{Lemma}
\newreptheorem{lem}{Lemma}

\usepackage{color}

\usepackage{hyperref}

\usepackage[accepted]{icml2016_arvix} 

\icmltitlerunning{Submission and Formatting Instructions for ICML 2016}
\icmltitlerunning{Normalized Hierarchical SVM}

\begin{document} 

\twocolumn[
\icmltitle{Normalized Hierarchical SVM}

\icmlauthor{Heejin Choi}{heejincs@ttic.edu}
\icmladdress{Toyota Technological Institute at Chicago,6045 S Kenwood Ave, Chicago, Illinois 60637}
\icmlauthor{Yutaka Sasaki}{yutaka.sasaki@toyota-ti.ac.jp}
\icmladdress{Toyota Technological Institute,2-12-1 Hisakata, Tempaku-ku, Nagoya, 468-8511, Japan}
\icmlauthor{Nathan Srebro}{nati@ttic.edu}
\icmladdress{Toyota Technological Institute at Chicago,6045 S Kenwood Ave, Chicago, Illinois 60637}

\icmlkeywords{boring formatting information, machine learning, ICML}

\vskip 0.3in
]

\begin{abstract}
We present improved methods of using structured SVMs in a large-scale
hierarchical classification problem, that is when labels are leaves,
or sets of leaves, in a tree or a DAG.  We examine the need to
normalize both the regularization and the margin and show how doing so
significantly improves performance, including allowing achieving
state-of-the-art results where unnormalized structured SVMs do not
perform better than flat models.  
We also describe a further extension
of hierarchical SVMs that highlight the connection between
hierarchical SVMs and matrix factorization models.
\end{abstract}

\newcommand{\fix}{\marginpar{FIX}}
\newcommand{\new}{\marginpar{NEW}}

\section{Introduction}

We consider the problem of hierarchical classification.  That is, a classification problem when the labels are leaves in a large hierarchy or
taxonomy specifying the relationship between labels.  Such hierarchies
have been extensively used to improve accuracy
  \cite{Mccallum98improvingtext,Silla:2011:SHC:1937796.1937884,Vural:2004:HMM:1015330.1015427}
in domains such as document categorization \cite{Cai:2004}, web content
classification \cite{Dumais:2000:HCW:345508.345593}, and image
annotation \cite{Huang:1998:AHI:290747.290774}.  In some problems,
taking advantage of the hierarchy is essential since each individual
labels (leaves in the hierarchy) might have only a few training
examples associated with it.

We focus on hierarchical SVM \cite{Cai:2004}, which is a structured SVM
problem with the structure specified by the given hierarchy.
Structured SVMs are simple compared to other hierarchical
classification methods, and yield convex optimization problems with
straight-forward gradients.  However, as we shall see, adapting
structured SVMs to large-scale hierarchical problems can be problematic and requires care.
We will demonstrate that ``standard'' hierarchical SVM suffers from
several deficiencies, mostly related to lack of normalization with
respect to different path-length and different label sizes in
multi-label problems, which might result in poor performance, possibly
not providing any improvement over a ``flat'' method which ignores the
hierarchy.  To amend these problems, we present the Normalized
Hierarchical SVM (NHSVM).  The NHSVM is based on normalization weights
which we set according to the hierarchy, but not based on the data.
We then go one step further and learn these normalization weights
discriminatively.  Beyond improved performance, this results in a
model that can be viewed as a constrained matrix factorization for
multi-class classification, and allows us to understand the
relationship between hierarchical SVMs and matrix-factorization based
multi-class learning  \cite{amit2007uncovering}.

We also extend hierarchical SVMs to issues frequently encountered in
practice, such as multi-label problems (each document might be labeled
with several leaves) and taxonomies that are DAGs rather then trees.

We present a scalable training approach and apply our methods to large
scale problems, with up to hundreds of thousands of labels and tens of
millions of instances, obtaining significant improvements over
standard hierarchical SVMs and state-of-the-art results on a
hierarchical classification benchmark.

\section{Related Work}

Much research was conducted regarding hierarchical multi-class or multi-label classification. The differences with   other methods lies in normalization of structure, scalability of the  optimization, and utilization of the existing label structure.

Our work is based upon hierarchical classification using SVM which is introduced in
 \citet{Cai:2004}. The model extends the multi-class SVM to hierarchical structure.   
        An extension to the multi-label case was presented
by  \citet{Cai_2007}. 
In  \citet{Rousu06kernel-basedlearning}, an efficient dual optimization method for a kernel-based structural SVM and weighted decomposable losses are presented for a tree structured multi-label  problem.
 These methods focus on dual optimization which does not scale up to our focused datasets with large instances and large number of labels.
  Also, the previous methods do not consider the normalization of the structures, 
 which is important for such large structures.

For instance, we focus on the Wikipedia dataset in Large Scale Hierarchical Text Classification Competition (LSHTC)\footnote{http://lshtc.iit.demokritos.gr/}. It has 400K instances with a bag of words representation of wikipedia pages which are multi-labeled to its categories. The labels are the leaves from  a DAG structure with 65K nodes. Notice that the scale of dataset is very large compared to dataset considered in previous mentioned methods. For instance, in \citet{Rousu06kernel-basedlearning} the largest dataset has 7K instances and 233 nodes. Extensions of KNN, meta-learning, and ensemble methods were popular methods in the competition.

\citet{gopal2013recursive} presented a model with a multi-task objective and an efficient parallelizable optimization method for dataset with a large structure and number of instances. However, its regularization suffers the same normalization issue, and relies on the other meta learning method\cite{gopal2010multilabel} in the post-processing for high accuracy in multi-label problems.

There are alternatives to SVMs approaches \cite{Weinberger_largemargin,Vural:2004:HMM:1015330.1015427,Cesa-Bianchi:2006:HCC:1143844.1143867}, however, the approaches are not scalable to large scale dataset with large structures.

 Another  direction is to learn the structure rather than utilizing given structure.  \citet{bravo2009estimating, blaschko2013taxonomic} focus on learning a small structure from the data, which is is very different from using a known structure. A fast ranking method\cite{prabhu2014fastxml} is proposed for a large dataset. It builds a tree  structure for ranking of labels. However, it does not utilize  given hierarchy, and is not directly a multi-label classifier.
 

%


\section{Preliminaries}

Let  $\mathcal{G}$ be a tree or a {\em directed acyclic graph} (DAG) representing a label
structure with  $M$ nodes. Denote the set of leaves  nodes in $\mathcal{G}$ as  $\mathcal{L}$.   For each $n\in [M]$,
define the sets of parent, children, ancestor, and descendent nodes of $n$ as $\mathcal{P}(n)$, 
$\mathcal{C}(n)$, 
  $\mathcal{A}(n)$,
  and  $\mathcal{D}(n)$,  respectively. Additionally, denote the ancestor nodes of $n$  including node $n$ as
  $\overline{\mathcal{A}}(n)=\{n\}\cup
  \mathcal{A}(n)$, and similarly, denote $\overline{\mathcal{D}}(n)$ for $\mathcal{D}(n)=\{n\}\cup\mathcal{D}(n)$. We also extend the notation
above for  sets of nodes to indicate the union of the corresponding
sets, i.e.,
$\mathcal{P}(A)=\cup_{n\in A}\mathcal{P}(n)$.

Let $\{ (x_i,y_i)\}^N_{i=1}$ be the training data of $N$ instances.
Each $x_i \in \mathbb{R}^d$ is a feature vector and it is labeled with
either a leaf (in single-label problems) or a set of leaves (in
multi-label problems) of $\mathcal{G}$.  We will represent the labels
$y_i$ as subsets of the nodes of the graph, where we include the
indicated leaves and all their ancestors.  That is, the label space
(set of possible labels) is $\mathcal{Y}_{s}=\{\overline{\mathcal{A}}(l)|l\in
\mathcal{L} \}$ for single-label problems, and 
$\mathcal{Y}_{m}=\{\overline{\mathcal{A}}(L)|L\subseteq
\mathcal{L}\}$ for multi-label problems.

\section{Hierarchical  Structured SVM}
We review the hierarchical structured SVM
introduced in  \citet{Cai:2004} and extended to the multi-label case in  \citet{Cai_2007}.
Consider $W\in\mathbb{R}^{M\times d}$, and let the $n$-th row vector $W_n$ be
be weights of the node $n\in [M]$.
 Define $\gamma(x,y)$ to be the  potential of label $y$ given feature $x$, which is the sum of the inner
products of $x$ with the weights of node $n\in y,$
$  \gamma(x,y)=\sum_{n \in y} W_n \cdot  x$.
If we vectorize $W$, $w=vec(W)=[W_1^T \; W_2^T\; \dots \; W_M^T]^T\in \mathbb{R}^{d\cdot M}$, and
define the class-attribute $\wedge(y)\in\mathbb{R}^M$, $[\wedge(y)]_n=1$ if $n\in y$
or 0 otherwise\footnote{The  class attributes could be variables, but only used  as a fixed constant for mathematical conveniences without detail discussions, and not used for normalization of the structure. }, then 
\begin{align} 
  \gamma(x,y)=\sum_{n \in y} W_n \cdot  x=w\cdot (\wedge(y)\otimes x) \label{classa}
\end{align} 
  where $\otimes$ is the Kronecker product. 
With weights $W_n$, prediction of an instance $x$ amounts to finding
the maximum response label 
\begin{align*}
\hat{y}(x)={\arg\max}_{y \in \mathcal{Y}} \gamma(x,y)={\arg\max}_{y \in \mathcal{Y}} \sum_{n\in y} W_n x \label{argmax}
\end{align*}
 Given a structural error $\triangle(y',y)$, for instance a hamming
 distance $\triangle^{H}(y',y)=|y'-y|= \sum_{n\in [M]}
 |\textbf{1}_{n\in y'}-\textbf{1}_{n\in y}|$, a training a
 hierarchical structured SVM is optimizing:
\begin{align}
\min_W\lambda &\underset{n}{\sum} \|W_n\|^2_2 \nonumber \\ 
&+\underset{i}{\sum}\max_{y\in\mathcal{Y}}\left \{ \sum_{n \in y}W_n x_i -\sum_{n \in y_i} W_n x_i  +\triangle(y,y_i)\right
\} 
\end{align}
Equivalently, in terms of $w$ and class-attribute $\wedge(y)$,
\begin{align}
&\min_W\lambda  \|w\|^2_2
+ \underset{i}{\sum}\max_{y\in\mathcal{Y}}\left \{ w\cdot ((\wedge(y)-\wedge(y_i))\otimes x_i) +\triangle(y,y_i)\right\}. \label{HSVM2}\end{align}

  \section{Normalized Hierarchical SVM}

\begin{figure}
\centering
      \includegraphics[width=.7\columnwidth]{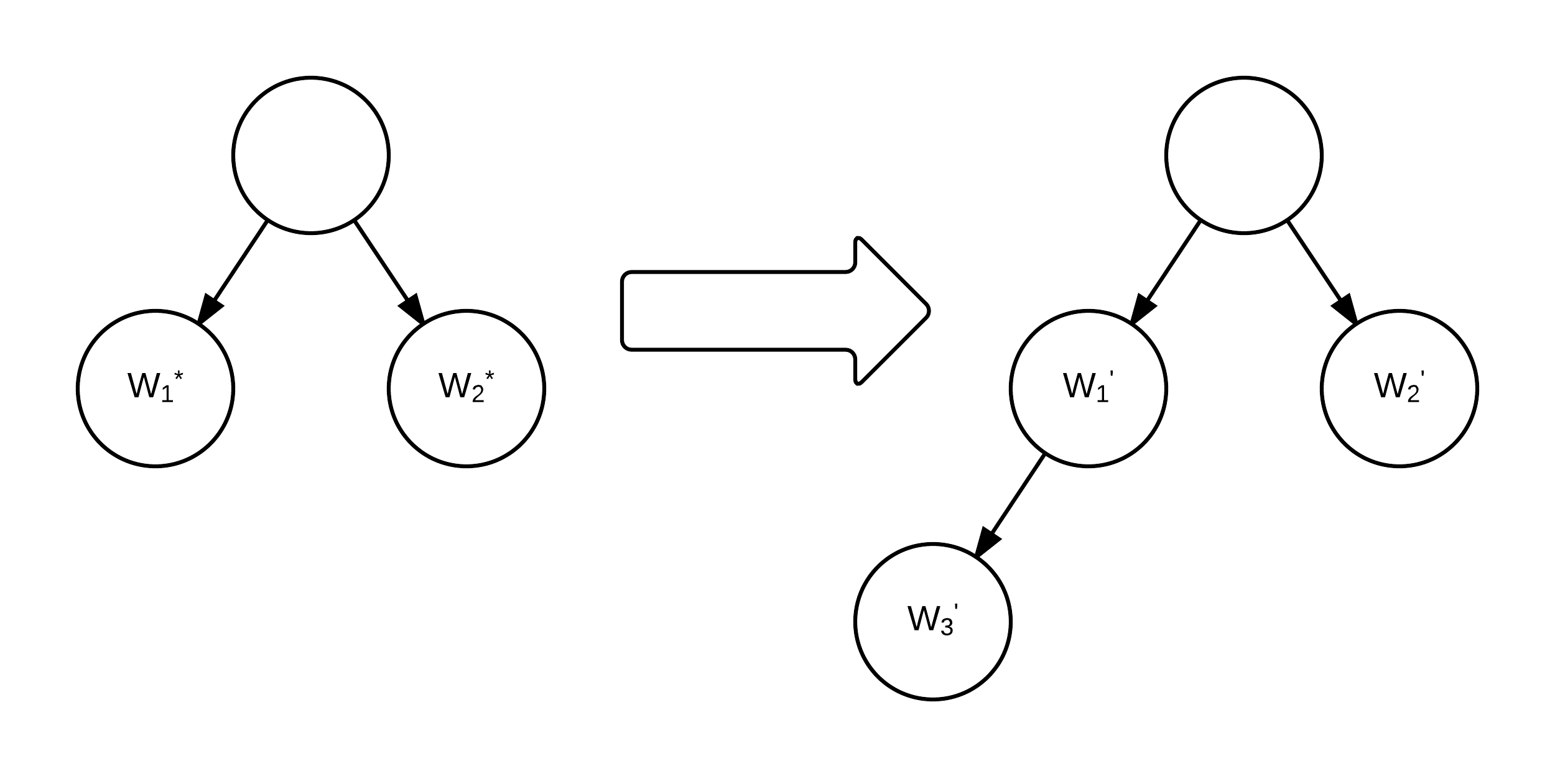}
    \caption{Regularization penalty for label $y_1$ (left branch) 
    is halved to 
    without
  changing decision boundary due to difference in the label structure. }
  \label{fig:nregularization}
\end{figure}


A major issue we highlight is that unbalanced structures (which are
frequently encountered in practice) lead to non-uniform regularization
with the standard hierarchical SVM.  To illustrate this issue,
consider the two binary tree structures with two leaves shown in figure
\ref{fig:nregularization}.  Implicitly both structures describes the
same structure. Recall that the regularization penalty is
$\|W\|^2_F=\sum_n\|W_n\|^2_F$ where each row of $W$ is a weight vector
for each node. In the left structure, the class attributes are $\wedge(y_1)=[1\; 0]^T$, and
$\wedge(y_2)=[0\; 1]^T$, assume $\|x\|_2=1$, and let the optimal
weights of node 1 and node 2 in the left structure be $W_1^*$ and
$W_2^*$. Now add a node 3 as a child of node 1, so that $M=3,
\wedge(y_1)=[1\; 0\; 1]^T,\wedge(y_2)=[0\; \; 1\; 0]^T$. Let $W_1'$
and $W_3'$ be the new weights for the nodes 1 and 3. If we assume
$W_1'=W_3'=\frac{1}{2}W^*_1$,  the potential function, and thus the
decision boundary remain the same, but the regularization penalty
for $y_1$ is halved so that 
$\|W_1'\|^2_2+\|W_3'\|^2_2=\frac{1}{2}\|W_1^*\|^2_2$, and
$\|W^*\|_F^2>\|W'\|_F^2$. This can be generalized to any depth, and
the regularization penalty can differ arbitrarily for the model with
the same decision boundary for different structures. In the given
example, the structure on the right imposes half the penalty for the
predictor of $y_1$ than that of $y_2$.
 
The issue can also be understood in terms of the difference between
the norms of $\wedge(y)$ for $y\in\mathcal{Y}$.
Let  $\phi(x,y)\in \mathbb{R}^{d\cdot
M}$ the feature map for an instance vector $x$ and a label $y$ such that $\gamma(x,y)=w\cdot \phi(x,y)$.
From (\ref{classa}),
\begin{align*}
w \cdot (\wedge(y)\otimes x)= w\cdot \phi(x,y)
\end{align*}
  
$\wedge(y)\otimes x$ behaves as a feature map in hierarchical structured SVM.
While the model regularizes $w$, the norm of $\phi(x,y)$ is different for $y$ and scales as $\|\wedge(y)\|_{2}$.
\begin{align*}
\|\gamma(x,y)\|_2=\|\wedge(y)\otimes x\|_2=\|\wedge(y)\|_2\cdot\|x\|_2
\end{align*}

Note that  $\|\wedge(y)\|_{2}=\sqrt{|\overline{\mathcal{A}}(y)|}$ 
and the differences in regularization can grow linearly with the depth
of the structure.
 
To remedy this effect, 
for each node $n$ we introduce a weight $\alpha_n\ge0$ 
   such that the sum of the weights along each path to a leaf is one, i.e.,
\begin{align}
 \sum_{n\in \overline{\mathcal{A}}(l)}\alpha_n=1, \;\;\;\;\;\; \forall l\in \mathcal{L}.\label{alpha}
\end{align}
Given such weights, we define  the normalized class-attribute $\tilde{\wedge}(y)\in \mathbb{R}^M$
 and  the normalized feature map $\tilde{\phi}(x,y)\in \mathbb{R}^{d\cdot M},$
\begin{align} \label{nclassa}
 &[\tilde{\wedge}(y)]_n=\begin{cases} \sqrt{\alpha_n} & \mbox{if } y\in n\\ 0 & \mbox{otherwise}\end{cases} &
&\tilde{\phi}(x,y)=\tilde{\wedge}(y)\otimes x 
\end{align}
The norm of these vectors are normalized to 1, independent
of  $y$, i.e., $\|\tilde{\wedge}(y)\|_2=1,\|\tilde{\phi}(x,y)\|_2=\|x\|_2$
for $y\in \mathcal{Y}_{s}$, and the class attribute for each node $n$ is fixed to
$0$ or $\sqrt{\alpha_n}$ for all labels.
The choice of $\alpha$ is crucial and we present several alternatives
(in our experiments, we choose between them using a hold-out set). For
instance, using $\alpha_n= 1$ on the leaves $n\in \mathcal{L}$ and 0
otherwise will recover the flat model and lose all the information in
the hierarchy.  To refrain from having a large number zero weight and
preserve the information in the hierarchy, we consider setting
$\alpha$ optimizing:
\begin{equation}
\begin{aligned}\label{beta_1}
&\min &&\sum \alpha_n^{\rho}\\ 
&\mbox{s.t.}& &\sum_{n\in \overline{\mathcal{A}}(l)}  \alpha_n=1, && \forall  l\in \mathcal{L}\\
&&&\alpha_n\ge0 &&\forall n\in [M]
\end{aligned}
\end{equation}
where $\rho>1$.  In Section \ref{sec:invar}, we will show that as
$\rho\rightarrow 1$, we obtain weights that remedy the effect of the
redundant nodes shown in
Figure \ref{fig:nregularization}.  

We use \eqref{beta_1} with $\rho=2$ as a possible way of setting the
weights.  However, when $\rho=1$, the optimization problem
\eqref{beta_1} is no longer strongly convex and it is possible to recover weights of zeroes for most nodes.  Instead, for $\rho=1$, we consider the alternative
optimization for selecting weights:
\begin{equation}
\begin{aligned}\label{nbeta_1}
&\max  & &\min_n\alpha_n&&\\ 
&\mbox{s.t.}& &\sum_{n\in \overline{\mathcal{A}}(l)  }  \alpha_n=1, \;\;\; &&\forall  l\in \mathcal{L}\\
&&&\alpha_n\ge0, &&\forall n\in [M] \\ 
&&&\alpha_n\ge\alpha_p,& &\forall n\in [M], \forall p\in \mathcal{P}(n)
\end{aligned}
\end{equation} 
We refer to the last constraint as a ``directional constraint'', as it
encourage more of the information to be carried by the leaves and results more even distribution of $\alpha$.

For some DAG structures, constraining the sum $\sum_{n\in
  \overline{\mathcal{A}}(l)}\alpha_n$ to be exactly one can
result in very flat solution.  For DAG structures we therefore relax
the constraint to
\begin{align}\label{con:range}
  1\le \sum_{n\in \overline{\mathcal{A}}(l) }  \alpha_n\le T, & &\forall  l\in \mathcal{L}.
 \end{align} 
for some parameter $T$ ($T=1.5$ in our experiments).

Another source of the imbalance is the non-uniformity of the required margin, which results from the norm of the differences of class-attributes, $\|\wedge(y)-\wedge(y')\|_2$.
The loss term of each instance in (\ref{HSVM2}) is,
$\max_{y\in \mathcal{Y}}w \cdot (\wedge(y)-\wedge(y_{i}))\otimes x+\triangle(y,y_i)$.
And to have a zero loss $\forall y\in \mathcal{Y}$,
\begin{align*}
\triangle(y,y_i)\le
w\cdot ((\wedge(y)-\wedge(y_{i}))\otimes x) 
\end{align*}
$\triangle(y,y_i)$ works as the margin requirement to have a zero loss for $y$. 
The RHS of the bound scales as norm of $\wedge(y)-\wedge(y_i)$
scales. 

%
%
%
 This calls for the use of structural
error that scales with  the bound. 
Define normalized structural
error $\tilde{\triangle}(y,y_i)$
\begin{align}
\tilde{\triangle}(y,y_i)=\|\tilde{\wedge}(y)-\tilde{\wedge}(y_i)\|=\sqrt{\sum_{n\in
y\triangle y_i} 
\alpha_n} \label{wserror}
\end{align}
and $y\triangle y'=(y_i-y) \cup (y-y_i)$
, and $\tilde{\wedge}(y)$
and
$\alpha$ are defined in (\ref{nclassa})(\ref{beta_1}).
Without the normalization, this is the square root of the hamming distance, and is similar to a tree induced distance in  \cite{dekel2004large}. This view of nonuniform margin gives a justification that the square root of hamming distance or tree induced
distance is preferable to hamming distance.
  
\subsection{Normalized Hierarchical SVM model}
 Summarizing the above discussion, we propose the Normalized
 Hierarchical SVM (NHSVM), which is given in terms of the following objective:
 \begin{align}
\min_W\lambda \underset{n}{\sum} \|W_n\|^2_2+\nonumber\\ \underset{i}{\sum}\max_{y\in\mathcal{Y}}  \sum_{n \in y}\sqrt{\alpha_n}W_n x_i -  
 &\sum_{n \in y_i} \sqrt{\alpha_n}W_n x_i  +\tilde{\triangle}(y,y_i)
 \label{NHSVM}
\end{align}
 Instead of imposing a weight for each node, with change of variables $U_n=\sqrt{\alpha_n}W_n$,
 we can write optimization (\ref{NHSVM}) as changing regularization, 
 \begin{align}\label{NHSVM2}
\min_U\lambda \underset{n}{\sum} \dfrac{\|U_n\|^2_2}{\alpha_n}\nonumber\\+ \underset{i}{\sum}\max_{y\in\mathcal{Y}}  \sum_{n \in y}U_n x_i -  
 &\sum_{n \in y_i} U_n x_i  +\tilde{\triangle}(y,y_i)
\end{align}
Also optimization \eqref{NHSVM} is equivalently written as 
 \begin{align}
\min_W\lambda  \|w\|^2_2+ \underset{i}{\sum}\max_{y\in\mathcal{Y}}w\cdot ((\tilde{\wedge}(y)-\tilde{\wedge}(y_i))\otimes x_i)+\tilde{\triangle}(y,y_i)
\end{align}
 Note that
for the single-label problem,  normalized hierarchical SVM can be viewed as a multi-class SVM changing the feature map function
  to  (\ref{nclassa}) and the loss term  to \eqref{wserror}. Therefore, it can be easily applied  to problems where flat SVM is used, and
also popular optimization method for SVM, such as  \citet{Shalev-Shwartz:2007:PPE:1273496.1273598, lacoste2013block},  can be used.

Another possible variant of optimization (\ref{NHSVM2}) which we experiment with is
obtained by dividing inside the max with
 $\|\tilde{\wedge}(y)-\tilde{\wedge}(y_i)\|_2$:
   \begin{align}
\min_W\lambda  \|w\|^2_2+ \underset{i}{\sum}\max_{y\in\mathcal{Y}}w\cdot \left (\frac{\tilde{\wedge}(y)-\tilde{\wedge}(y_i)}{\|\tilde{\wedge}(y)-\tilde{\wedge}(y_i)\|_2}\otimes x_i\right )+1
 \label{NHSVM3}
\end{align}
 There are two interesting properties of the optimization \eqref{NHSVM3}. The norm of the vector right side of $w$ is normalized, i.e.,  
\begin{align*}
\left\|\frac{\tilde{\wedge}(y)-\tilde{\wedge}(y_i)}{\|\tilde{\wedge}(y)-\tilde{\wedge}(y_i)\|_2}\otimes x_i\right \|_2=\|x_i\|_2.
\end{align*}
 Also the loss term per instance at the decision boundary, which is also the required margin, is normalized to 1. However, because normalized class attribute in (\ref{NHSVM3}) does not decompose w.r.t nodes as in (\ref{NHSVM}), 
loss augmented
inference in (\ref{NHSVM3}) is not efficient for multi-label problems.

\subsection{Invariance property of the normalized hierarchical SVM}\label{sec:invar}

As we saw in Figure \ref{fig:nregularization}, different hierarchical structures  can be used to  describe the same data, 
and this causes undesired regularization problems. However, this is a common problem in real-world datasets. For
instance, an {\em action } movie label can be further categorized into a {\em
cop-action} movie and a {\em
hero-action} movie in one dataset whereas the other dataset uses a action movie as a label. 
Therefore, it is desired for the learning method of
hierarchical
model to adapt to this difference and learn a similar model if given dataset describes
similar  data.
 Proposed normalization can be viewed as an adaptation to this kind of distortions. In particular,  we show that NHSVM is invariant to   node duplication.  

 Define duplicated nodes as follows. 
Assume that there are no unseen nodes in the dataset, i.e., $\forall n\in\mathcal{[M]},\exists
i, n\in \mathcal{A}(y_i)$.
Define two  nodes  $n_1$ and $n_2$ in $[M]$ to be {\em duplicated}  if $\forall i, 
 n_1\in y_i\iff n_2 \in y_i$.
Define the minimal graph $M(\mathcal{G})$ to
be  the graph having  a representative node per each duplicated
node set by merging each duplicated
node set to a node. For the proof, see \ref{apx:invariance}.

\begin{thm}[Invariance property of NHSVM] \label{thm:invarinace}
Decision
boundary of NHSVM with $\mathcal{G}$ is arbitrarily close to that   of NHSVM with the minimum graph $M(\mathcal{G})$ as $\rho$ in (\ref{beta_1}) approaches 1, $\rho>1$.

\end{thm}

\section{Shared SVM: Learning with Shared Frobenius norm}

In the NHSVM, we set the weights $\alpha$ based the graphical structure
of the hierarchy, but disregard the data itself.  We presented
several options for setting the weights, but it is not clear what the
best setting would be, or whether a different setting altogether would
be preferable.  Instead, here we consider discriminative learning the
weights from the data by optimizing a joint objective over the weights
and the predictors.  The resulting optimization is equivalent to  regularization with a new norm which we  call {\em Structured Shared Frobenius norm}  or {\em Structured Shared norm}. It explicitly  incorporates the information of  the label structure $\mathcal{G}$. Regularization with the structured shared Frobenius norm promotes the models to utilize shared information, thus it  is a complexity measure  suitable for  structured learning. Notice that we only consider multi-class problem in this section. An efficient algorithm for tree structure is discussed in section \ref{sec:opt}.

Consider the
formulation \eqref{NHSVM2} as a joint optimization over both $\alpha$ and
$U=[U_1^T\; U_2^T\; \dots U_M^T]^T$ with fixed
$\tilde{\triangle}(y,y_i)=\triangle(l,l_i)$ (i.e.~we no longer
normalize the margins, only the regularization):
\begin{equation} 
\begin{aligned} \label{SSVMp}
&\min_{U,\alpha}& \lambda \underset{n}{\sum} \dfrac{\|U_n\|^2_2}{\alpha_n} + &\underset{i}{\sum}\max_{l\in[Y]} \left \{ \sum_{n \in \overline{\mathcal{A}}(l)}U_n x_i \right . -  \\
 && &\left . \sum_{n \in \overline{\mathcal{A}}(l_i)} U_n x_i  +\triangle(l,l_i) \right \}\\
 &\mbox{s.t.} &\sum_{n\in \overline{\mathcal{A}}(l)}  \alpha_n&\le1, & \forall  l\in [Y]\\
 &&\alpha_n&\ge0 &\forall n\in [M]
\end{aligned}
\end{equation}
We can think of the first term as a regularization norm
$\|\cdot\|_{s,\mathcal{G}}$ and write
\begin{align} \label{ssvm}
&\min_U\lambda  \|U\|^2_{s,\mathcal{G}}+ 
\underset{i}{\sum}\max_{l\in |Y|}U_l\cdot x_i - U_{l_i}\cdot x_i  +\triangle(l,l_i) \end{align}
where the the {\em structured shared Frobenius norm} $\|\cdot\|_{s,\mathcal{G}}$ is defined as:
\begin{equation}\label{rho2}
\begin{aligned}
 \|U\|_{s,\mathcal{G}}&=\min_{ a\in\mathbb{R}^M,V\in \mathbb{R}^{M\times d}} \|A\|_{2\rightarrow\infty}\|V\|_F   \\
\mbox{s.t. }  AV&=U\\ 
 A_{l,n}&= \begin{cases}0 & otherwise\\ 
 a_n
& n\in \overline{\mathcal{A}}(l)\end{cases}  &\forall l, \forall n\\
 a_n&\ge0,&\forall n\in [M]
\end{aligned}
\end{equation}
where $\|A\|_{2\rightarrow \infty}$ is the maximum of the $\ell_2$ norm of
row vectors of $A$.  Row vectors of $A$ can be viewed as coefficient
vectors, and row vectors of $V$ as factor vectors which decompose the
matrix $U$.  The factorization is constrained, though, and must
represent the prescribed hierarchy.  We will refer (\ref{ssvm}) to
{\em Shared SVM} or {\em SSVM}.

To better understand the SSVM, we can also define the {\em Shared Frobenius norm}  without the structural constraint as 
\begin{align}
 \|U\|_{s}=\min_{ AV=U} \|A\|_{2\rightarrow \infty}\|V\|_F   \label{eq:snorm1}
\end{align}
The {\em Shared Frobenius norm} is a norm between the trace-norm (aka
nuclear norm) and the max-norm (aka $\gamma_2:1\rightarrow\infty$
norm), and an upper bounded by Frobenius norm:
\begin{thm}\label{thm:compare}
For $\forall U\in \mathbb{R}^{r\times c}$
\begin{align*}
 \dfrac{1}{\sqrt{rc}}\|U\|_* &\le\dfrac{1}{\sqrt{c}}\|U\|_s\le\|U\|_{\max}\\
 \|U\|_{s}&\le\|U\|_{s,\mathcal{G}}\le \|U\|_F
\end{align*}
where $\|U\|_*=\min_{AW^T=U}\|A\|_F\|W\|_F$ is then the trace norm, and $\|U\|_{\max}=$
$\min_{AW^T=U}$ $\|A\|_{2\rightarrow
\infty}\|W\|_{2\rightarrow
\infty}$ is  so-called the max norm \cite{srebro2005rank}.

\proof The first  inequality follows from the fact that $\frac{1}{\sqrt{r}}\|U\|_{F}\le \|U\|_{2\rightarrow\infty}$,
and the second  inequality  is from taking $A=I$, or $A_{l,n}=1$ when $n$ is  an
unique node for $l$ or 0 for all other nodes in (\ref{rho2}) respectively.  \qed
\end{thm}

We compare the Shared norm to the other norms to illustrate the behavior
of the Shared norm, and summarize in Table \ref{compare_norms}.  Shared norm is upper bounded by Frobenius norm, and reduce from it only if sharing the factors $V$ is beneficial. If there is no reduction from sharing as in disjoint feature case  in Table \ref{compare_norms}, it equals to  Frobenius norm,  which is the norm used for multi-class SVM.   Therefore, this  justifies the view of SSVM that it extends multi-class SVM
to shared structure, i.e., SSVM is equivalent to multi-class SVM if no sharing of weights is beneficial. This differs from the trace norm, which we can see  specifically in disjoint feature case. 
    
\begin{table*}[t]
\centering
\begin{tabular}{ |l|c | c | c| c| }
\hline
   & $\|U\|_{s}$& $\|U\|_{s,\mathcal{G}}$ & $\|U \|_F$ & $\|U\|_*$\\
\hline
  Full sharing & $\|u\|_2$ & $\|u\|_2$ & $\sqrt{Y}\|u\|_2$ & $\sqrt{Y}\|u\|_2$\\
\hline
  No sharing  & .& $\|U\|_F$ & $\|U\|_F$ & $\cdot$ \\
  \hline
  Disjoint feature & $\sqrt{\sum_l \|u_l\|_2^2}$& $\sqrt{\sum_l \|u_l\|_2^2}$ & $\sqrt{\sum_l \|u_l\|^2_2}$ & $\sum_l \|u_l\|_2$ \\
  \hline
  Factor scaling & $\max_i |a_i|{\|u\|_2}$ & . &  $\sqrt{\sum_i a_i^2}\|u\|_2$ & $\sqrt{\sum_i a_i^2}\|u\|_2$ \\
  \hline\end{tabular}
  \caption{ Comparing $\|U\|_{s}, \|U\|_{s,\mathcal{G}}, \|U\|_F$ and
  $\|U\|_*$
in different situations. 
(1) Full sharing,
$U=[u\; u\;
\dots\; u]^T,\exists n',\forall l,n'\in \bar{\mathcal{A}}(l)$. (2) No sharing,  $\forall l\neq l', \bar{\mathcal{A}}(l)\cap \bar{\mathcal{A}}(l')=\emptyset$.
(3) Disjoint  feature, 
$U=[u_{1}\; u_{2}\;
\dots \; u_Y]^T, \forall l_1\neq l_2, \mbox{Supp}(u_{l_1})\cap \mbox{Supp}(u_{l_2})=\emptyset$. (4) Factor scaling, $U=[a_1 u\; a_2 u
\dots \; a_Y u]$.
Unlike trace norm, Shared Frobenius norm  reduces to Frobenius norm if no sharing is beneficial  as in case of (3)\ disjoint feature.  See the text and \ref{apx:shared} 
for   details.
}\label{compare_norms}
\end{table*}  
\section{Optimization} \label{sec:opt}
 In this section, we discuss the details of optimizing objectives \eqref{NHSVM} and \eqref{SSVMp}.
 Specifically, we show how to obtain the most violating label for multi-labels problems for  objective \eqref{NHSVM} and an efficient algorithm to optimize objective \eqref{SSVMp}.  
\subsection{Calculating the most violating label for multi-label problems}

We optimize our training objective \eqref{NHSVM} using SGD \cite{Shalev-Shwartz:2007:PPE:1273496.1273598}.
 In order to do so, the most challenging part is calculating
\begin{align} \label{argmax_}
\hat{y}_{i}=\arg\max_{y\in\mathcal{Y}}  \sum_{n \in y}\sqrt{\alpha_n}W_n x_i -  
 &\sum_{n \in y_i} \sqrt{\alpha_n}W_n x_i  +\tilde{\triangle}(y,y_i)\nonumber\\=\arg\max_{y\in\mathcal{Y}}
 L_{i}(y)
 \end{align}
 at each iteration. For single label problems, we can calculate $\hat{y}_i$ by enumerating all the labels. However, for a multi-label problem, this is intractable
because of the exponential size of the label set. Therefore, in this subsection,  we describe how to calculate $\hat{y}_i$ for  multi-label problems. 

If $L_i(y)$ decomposes as a sum of functions with respect to its nodes, i.e., $L_{i}(y)=\sum_n L_{i,n}(\text{1}\{n\in y\})$, then 
$\hat{y}_i$ can be found efficiently. 
Unfortunately,  
 $\tilde{\triangle}(y,y_i)$ does not decompose with respect to the nodes.   In order to allow efficient computation for  multi-label problems, we actually replace $\tilde{\triangle}(y,y_i)$ with a decomposing approximation  $\triangle_\mathcal{L} l(y,y_i)=|\{y\cap \mathcal{L}\}-\{ y_{i}\cap \mathcal{L}\}|$
$=\sum_{l\in\mathcal{L}}\left ({\bf{1}}_{\{l\in y- y_{i}\}}-{\bf{1}}_{\{l\in y\cap y_{i}\}}\right)+|\{ y_i\cap \mathcal{L}\}|$ instead. When $\triangle_\mathcal{L} l(y,y_i)$ is used and the  graph $\mathcal{G}$ is a tree,   $\hat{y}_i$ can be computed in time $O(M)$ using dynamic programming. 


When the graph $\mathcal{G}$ is a DAG,  dynamic programming is not applicable.
However, finding \eqref{argmax_} in a DAG structure can be formulated
into  the following integer programming.  
\begin{align}
\begin{aligned}
&\hat{z}=&\underset{ z\in \{0,1\}^M}{\arg\min} &\sum_{n=1}^M z_n \cdot  r_n \\
&\mbox{s.t} 
  &\sum_{c \in \mathcal{C}(n)}z_c &\ge z_n,  &&\forall n \\
& &z_c &\le z_n,  && \forall n, \forall c \in \mathcal{C}(n) \\
& & \sum_{l\in \mathcal{L}}z_l&\ge1  
\end{aligned}\label{integerprogram}
\end{align}
where $r_n=\sqrt{\alpha_n}W_n x_i +{\bf
{1}}_{\{n\notin  y_{i},n\in\mathcal{L}\}}-{\bf
{1}}_{\{n\in y_{i},n\in\mathcal{L}\}}$. The feasible label  from \eqref{integerprogram} is the set of labels where if a node $n$ is in the label $y$, at least one of its child node is in $y$, i.e.,  $\forall n\notin \mathcal{L}, n\in y \implies \exists c\in\mathcal{C}(n), c\in y$, and all the parents of $n$ are in the label, i.e., $\forall n\in y\implies \forall p\in \mathcal{P}(n), p\in y$. The feasible set  is equivalent to $\mathcal{Y}_m$.
 The search problem (\ref{integerprogram}) can be shown to be NP-hard by reduction from the set cover problem. We relax the  integer program into a linear program for training. Last constraint of $\sum_{l\in \mathcal{L}}z_l\ge1$ is not needed for an integer program, but yields a tighter LP relaxation. In testing, we rely on the binary integer programming only if the solution to
LP is not integral. In practice, integer programming solver is effective for this problem, only 3 to 7
times slower than linear relaxed program using gurobi solver \cite{gurobi}.

\subsection{Optimizing with the shared norm}
Optimization \eqref{SSVMp} is a convex  optimization jointly  in  $U$ and $\alpha$, and thus,  has a global optimum. 
%
%
For the proof, see \ref{apx:convex}.
\begin{lem}\label{lem:convex}
Optimization \eqref{SSVMp} 
is a convex optimization jointly in $U$ and $\alpha$. 
\end{lem}
Since it is not clear how to  jointly optimize efficiently with respect to $U$ and $\alpha$,  we present an efficient method to optimize \eqref{SSVMp} alternating between $\alpha$ and $V_{n}=\frac{U_n}{\sqrt{\alpha_n}}$. Specifically, we show how to calculate the optimal $\alpha$ for a fixed $U$ in closed form in time $O(M)$ when  $\mathcal{G}$ is a tree where $M$ is the number of nodes in the graph, and for fixed $\alpha$, we optimize the objective using SGD \cite{Shalev-Shwartz:2007:PPE:1273496.1273598} with change of variables $V_{n}=\frac{U_n}{\sqrt{\alpha_n}}$.
%
\begin{align}\label{optimize_alpha}
&\min_{\alpha_n}&
 \sum_{n\in [M]}\dfrac{\|U_n\|_2^2}{\alpha_n}&
\\
&\mbox{s.t}&\sum_{n\in \mathcal{A}(y)} \alpha_n\le 1,&&\forall y\in \mathcal{Y} \nonumber
\\
& &\alpha_n\ge 0,&&\forall n\in \mathcal{N} \nonumber.
\end{align}

\begin{algorithm}[tb]
 \caption{Calculate the optimal $\alpha$ in \eqref{optimize_alpha} for the tree structure $\mathcal{G}$ in $O(M)$.  We assume nodes are sorted in increasing order of depth,i.e., $\forall n, p\in \mathcal{P}(n), n>p$.
}\label{argmin_alpha}
 \begin{algorithmic}[1]
   \STATE {\bfseries Input:} {$U\in \mathbb{R}^{M\times d},$ a tree graph $\mathcal{G}$}
   \STATE {\bfseries Output:}{ $\alpha \in \mathbb{R}^{M}$.}
   \STATE {\bfseries Initialize:}{ $\alpha=N=E=[0\;0\dots\;0] \in\mathbb{R}^M, L=[1 \;1\dots1]\in\mathbb{R}^M$}
   \FOR {$n = M \to 1$} 
       \STATE $N_n\gets \|U_{n,\cdot}\|_{2}^2,E_n\gets 0$
     \IF{$|\mathcal{C}(n)|\neq0$} 
     \STATE $S\gets \sum_{c\in\mathcal{C}(n)}N_c$
       \IF{$\sqrt{S}+\sqrt{N_n}\neq0$} 
       \STATE $N_n \gets (\sqrt{N_n}+\sqrt{S})^2,E_n\gets \dfrac{\sqrt{N_n}}{\sqrt{S}+\sqrt{N_n}}$
       \ENDIF{}
     \ENDIF{}
   \ENDFOR
   \STATE $L_1\gets 1- E_1$
   \FOR {$n = 2 \to M$} 
       \STATE $p\gets \mathcal{P}(n)$
       \STATE $\alpha_n=\sqrt{L_{p}\cdot E_n}$
       \STATE $L_n=L_{p}\cdot(1-E_n)$
   \ENDFOR
     \STATE \textbf{return} $\alpha$
 \end{algorithmic}
\end{algorithm}
Algorithm \ref{argmin_alpha} shows how to calculate optimum $\alpha$ in  \eqref{optimize_alpha} in time $O(M)$ for a tree structure. See \ref{apx:tree_shared}.
\begin{lem}
\label{lem:tree_shared} For a tree structure $\mathcal{G}$, algorithm \ref{argmin_alpha} finds optimal $\alpha$ in \eqref{optimize_alpha} in $O(M)$ in a closed form.
\end{lem}
In the experiments, we optimize $\alpha$ using algorithm \ref{argmin_alpha} with $U_n=\sqrt{\alpha_n} V_n$ after a fixed number of epochs of SGD with respect to $V$, and repeat this until the objective function converges. We find that the algorithm is efficient enough to scale up to large datasets.

\section{Experiments}

We present experiments on both synthetic and real data sets.  In Section \ref{sec:syn}, we consider synthetic data sets with both balanced structures and unbalanced structures (i.e. when some leaves in the class hierarchy are much deeper then others).  We use this to demonstrate empirically the vulnerability of un-normalized Hierarchical SVM to structure imbalance, and how normalization solves this problem.  In particular, we will see how un-normalized HSVM does not achieve any performance gains over "flat" learning (completely ignoring the structure), but our NHSVM model does leverage the structure and achieves much higher accuracy.  Then, in Section \ref{sec:bench}, we compare our method to  competing methods on mid-sized benchmark data sets, including ones with multiple labels per instance and with DAG structured hierarchies.  Finally, in Section \ref{sec:comp} we demonstrate performance on the large-scale LSHTC competition data, showing significant gains over the previously best published results and over other recently suggested methods.  Data statistics
 is summarized in table \ref{exper_spec}.  

\begin{table}
\centering
\begin{tabular}{| l | c | c |c|c|c|}
    \hline
     & $M$& $d$&$N$&$|\mathcal{L}|$& $\overline{\mathcal{L}}$\\
    \hline
    Synthetic(B)  & 15 & 1K& 8K & 8 & 1\\
    Synthetic(U)  & 19 & 1K& 10K & 11 & 1\\
        IPC & 553& 228K& 75K& 451 & 1
\\   
    WIKI d5 & 1512& 1000&41K&1218 & 1.1\\
    ImageNet & 1676&51K&100K&1000 & 1\\
    DMOZ10  & 17221 & 165K& 163K & 12294 & 1\\
    WIKI  & 50312& 346K&456K&36504 & 1.8\\
    \hline
    \end{tabular}
  \caption{Data statistics: $M$ is the number of nodes in the graph. $d$
  is the dimension of the features. $N$ is the number of the instances.  $|\mathcal{L}|$
  is the number of labels.
 $\overline{\mathcal{L}}$ is the average labels per instance. $\overline{\mathcal{L}}=1$
 denotes a single-label dataset.
 }\label{exper_spec}
\end{table}

\subsection{Synthetic Dataset}\label{sec:syn}
In this subsection, we empirically demonstrate the benefit of the normalization  with the intuitive hierarchical synthetic datasets. While even for a perfectly balanced structure, we gain from the normalization, we show that the regularization of the HSVM can suffer  significantly from imbalance of the structure (i.e. when the depths of the leaves are very different). Notice that for a large structured dataset such as wikipedia dataset, the structure is very unbalanced.

Balanced synthetic data is created as follows. A weight vector $W_n\in\mathbb{R}^d$ for each node $n\in[2^3-1]$ in the complete balanced tree with depth 4 and an instance vector $x_i\in\mathbb{R}^d$,$i\in [N]$,$N=15000,$ for each instances are sampled from the standard multivariate normal distribution. Instances are assigned
to labels which have maximum potential.
To create the unbalanced synthetic data,
 we sample $x_i\in \mathbb{R}^d$ from the multivariate normal distribution
 with $d=1,000, i\in [N], N=10,000$, and normalize its norm to 1.  We divide
 the space $\mathbb{R}^d$ with a random hyperplane recursively so that the divided spaces form an unbalanced binary tree structure, a binary tree growing only
in one direction.  Specifically, we divide the space  into two spaces with a random hyperplane, which form two child
spaces, and recursively
divide only one of the child space with a random hyperplane until the depth of the binary tree reaches 10. 
Each $x$ is assigned to leaf nodes if $x$ falls into the corresponding space. 

In both datasets, our proposed models, NHSVM and SSVM, are compared to HSVM \cite{Cai:2004}, and flat SVM in the Table \ref{syn}. 
For each experiments the different parameters are tested on the the holdout dataset. Fixed set of $\lambda$ is tested, $\lambda\in\{10^{-8},10^{-7},\dots,10^{2}\}$. For NHSVM is tested with $\rho=2$, and $\rho=1$  in \eqref{NHSVM} and \eqref{NHSVM3}. Also $\rho=2$ is tested with directional constraints. For both WIKI, $T=1.5$ is used in \eqref{con:range}.
And each model with the parameters which had the best holdout error  is trained
 with all the training data, and we report test errors. We repeated the test for 20 times, and report the mean and the standard deviations. Notice that HSVM fails to exploit the hierarchical structure of the unbalanced dataset with the accuracy less than flat model, whereas NHSVM achieves higher accuracy by 6\% over flat model. The accuracy gain of NHSVM against HSVM for the balanced dataset, shows the advantage of (\ref{NHSVM2}) and normalized structured loss(\ref{wserror}). For the unbalanced dataset, SSVM further achieves around 1\% higher accuracy compared to NHSVM learning the underlying structure from the data.  For the balanced dataset, SSVM performs similar to NHSVM.

\begin{table}
\centering
 \begin{tabular}{| c | c | c|c|}

    \hline
    Method & Balanced& Unbalanced \\
    \hline
     SSVM & 63.4 $\pm$.35& 74.9$\pm$.4$^\ddagger $\\
     NHSVM &63.3 $\pm$.34$^\dagger$& 74.1$\pm$.2$^\dagger$\\     
     HSVM &62.8$\pm$.39   & 68.4$\pm$.07 \\
     FlatSVM  &60$\pm$.24 \ & 68.5$\pm$.1\\
    \hline  
    \end{tabular}
  \caption{Accuracy on synthetic datasets. $\dagger$ shows that the  the improvements over FlatSVM and HSVM is statistically significant. $\ddagger$ shows that the improvement over NHSVM is statistically significant.  }\label{syn}
\end{table}

\subsection{Benchmark Datasets}\label{sec:bench}

We show the benefit of our model on  several real world benchmark datasets in different fields without
restricting domain to the document classification, such as  ImageNet in table
\ref{bench_result}. We followed same procedure described in section \ref{sec:syn}. Results  show  consistent improvements
over our base models. NHSVM outperforms our base methods, and
 SSVM shows additional increases in the performance.
DMOZ 2010 and WIKI-2011 are
from LSHTC competition.  
IPC\footnote{http://www.wipo.int/classifications/ipc/} is a single label patent document dataset. DMOZ 2010 is a  single label web-page collection. 
WIKI-2011
is a multi-label dataset of wikipedia pages, depth is cut to 5 (excluded
labels with depth more than 5). 
ImageNet data \cite{ILSVRCarxiv14} is a single label image data with
SIFT BOW features from development kit 2010.
WIKI and ImageNet have DAG structures, and the others have tree structures.
\begin{table}
\centering
    \begin{tabular}{| l | c |c|c|c| }
    \hline
      Method & IPC 
      &DMOZ 
      &  WIKI d5 & Imagenet \\
    \hline
%
      SSVM &  52.6$\pm$.069$^\ddagger$     
      &45.5 
      & *      & *\\ 
      NHSVM & 52.2$\pm$.05$^\dagger$
      &45.5 
      & 60$\pm$.87$^\dagger$ &8.0$\pm$.1$^\dagger$\\
      HSVM     & 50.4$\pm$.09  
      &45.0 
      & 58$\pm$1.1 & 7.3$\pm$.16\\
      FlatSVM & 51.6$\pm$.08\ 
      & 44.2 
      & 57$\pm$1.3 & 7.6$\pm$.08\\
    \hline
    
    \end{tabular}
  \caption{Accuracy on benchmark datasets. * denotes that the algorithm was not able to be applied due to the graphical structure of the data.
}
\label{bench_result}
\end{table}
\subsection{Result on LSHTC Competition }\label{sec:comp}
We also compared our methods with the competition dataset, Large Scale Hierarchical Text Classification Challenge 2\footnote{http://lshtc.iit.demokritos.gr/}. We compared with the winner of the competition as well as the the best published method we acknowledge so far, HR-SVM \cite{gopal2013recursive}. 
We also added comparisons with
FastXML\cite{prabhu2014fastxml} in the competition dataset. FastXML is a very fast ranking method suitable for a large dataset. Since FastXML  predicts rankings of full labels rather than list of labels, we predicted with the same number of labels as NHSVM, and compared the result. 
 In table \ref{comp_result}, we show the result on full competition dataset, WIKI-2011, and compare with results currently reported. NHSVM was
 able to adapt to the large scale of WIKI-2011 dataset with the state-of-the-art
 results. Only 98,519 features that appear in the test set are used with tf-idf type weighting BM25 \cite{robertson2009probabilistic}. With a computer with  Intel Xeon CPU E5-2620   processor, optimization took around 1.5 weeks in matlab without a warm start.
\begin{table}
    \begin{tabular}{ |c | c |}
    \hline
      Method & Accuracy \\
    \hline
      NHSVM     &  43.8 \\
      HSVM      &  41.2 \\
       HR-SVM*\cite{gopal2013recursive}\    & 41.79 \\
       FastXML**\cite{prabhu2014fastxml}\ & 31.6\\
      Competition Winner    & 37.39  \\
    \hline
    \end{tabular}
  \caption{Results on full WIKI. *The inference of HR-SVM relies on the other meta learning method\cite{gopal2010multilabel} for high accuracy. ** NHSVM is used to predict the number of labels in the inference.   }\label{comp_result}
\end{table} 

\section{Summary}
In this paper we considered the problem of large-scale hierarchical
classification, with a given known hierarchy.  Our starting point was
hierarchical structured SVM of \citet{Cai:2004}, and we also
considered extensions for handling multi-label problems (where each
instance could be tagged with multiple labels from the hierarchy) and
of label hierarchies given by DAGs, rather then rooted trees, over the
labels.  Our main contribution was pointing out a normalization
problem with this framework, both in the effective regularization for
labels of different depths, and in the loss associated with different
length paths.  We suggested a practical correction and showed how it
yields to significant improvement in prediction accuracy.  In fact, we
demonstrate how on a variety of large-scale hierarchical
classification tasks, including the Large-scale Hierarchical Text Classification Competition data, our
Normalized Hierarchical SVMs outperform all other relevant methods we
are aware of (that work using the same data and can be scaled to the
data set sizes).  We also briefly discussed connections with matrix
factorization approaches to multi-label classification and plan on
 investigating this direction further in future research.
%
 

\small{

\bibliographystyle{apalike}
\bibliography{NHSVM_arvix}

}

\clearpage
\appendix
\gdef\thesection{Appendix \Alph{section}}
\section{Invariance property of NHSVM}
\label{apx:invariance}

\begin{repthm}{thm:invarinace}[Invariance property of NHSVM]
Decision
boundary of NHSVM with $\mathcal{G}$ is arbitrarily close to that   of NHSVM with the minimum graph $M(\mathcal{G})$ as $\rho$ in (\ref{beta_1}) approaches 1, $\rho>1$. 
\proof 
We prove by showing that  for any $\mathcal{G,}$ variable $\alpha$ in (\ref{beta_1}) can
be reduced to one variable per each set of  duplicated nodes in
$\mathcal{G}$ using the optimality conditions, and optimizations    (\ref{beta_1})(\ref{NHSVM}) are equivalent to the corresponding optimizations of  $M(\mathcal{G})$  by  change of
the variables .

Assume there are no duplicated leaves, however, the proof can be easily generalized
for the duplicated leaves by introducing an additional 
constraint on $\mathcal{Y}$. 

Let $\mathcal{F}(n')$ be a mapping from node $n'$ in graph $M(\mathcal{G})$ to a corresponding  set of duplicated nodes in $\mathcal{G}$.
Denote the set of nodes in $\mathcal{G}$ as $\mathcal{N}$, and the set of nodes in $M(\mathcal{G})$ as $\mathcal{N}'$,
 and the set of leaves in $M(\mathcal{G})$ as $\mathcal{L}'$.


Consider (\ref{beta_1}) for $\mathcal{G}$. Note that (\ref{beta_1}) has a constraint on sum of $\alpha_{n}$ to be 1 for $n\in\{n\in \bar{\mathcal{A}}(l)| l\subseteq\mathcal{L}\}$. By the
 definition of the  duplicity, if two nodes $n_1$ and $n_2$ are duplicated nodes,
they are the ancestors of the same set of the leaves, and term $\alpha_{n_1}$ appears in the first constraints of (\ref{beta_1}) if and only if term $\alpha_{n_2}$ appears, thus we conclude that all the duplicated nodes will appear altogether. Consider a change
of variable for each $n'\in \mathcal{N}'$ 
\begin{align}
K_{n'}=\sum_{n \in \mathcal{F}(n')}\alpha_{n}  \label{sub_k}
\end{align}
  Then, (\ref{beta_1})   are functions of $K_{n'}$ and (\ref{beta_1})
 decompose w.r.t $K_{n'}$. From the convexity of function $x^\rho$ with $\rho>1$,
  $x>0$,
 and Jensen's inequality, $(\frac{1}{|\mathcal{F}(n')|}K_{n'})^\rho\le\frac{1}{|\mathcal{F}(n')|}\sum_{n\in
 \mathcal{F}(n')}\alpha_n^\rho$,
 minimum of (\ref{beta_1}) is attained when $\alpha_{n}= \frac{1}{|\mathcal{F}(n')|}K_{n'}$ for $\forall n\in \mathcal{F}(n')$. As $\epsilon$ approaches  0, where\ $\epsilon=\rho-1>0$,
\begin{align}
\sum_{n\in\mathcal{N} } \alpha_{n}^{\rho }=\sum_{n'\in \mathcal{N}'} |\mathcal{F}(n')  |\left(
 \dfrac{K_{n'}}{|\mathcal{F}(n')|}\right )^{\rho} =|\mathcal{F}(n')|^{\epsilon}  K_{n'}^{\rho} \label{KK}
\end{align}

Plugging (\ref{KK}) (\ref{sub_k}) into (\ref{beta_1}), 
\begin{align*}
&\min&& \sum_{n'\in \mathcal{Y}'} K_{n'}^{\rho}\\
&\mbox{s.t.}&& \sum_{n'\in y'} K_{n'}=1, && \forall\ y'\in \mathcal{Y}'
\end{align*}
 These formulations are same as (\ref{beta_1}) 
 for $M(\mathcal{G})$.
 
  Thus  given $n'$,  $\alpha_{n}= \frac{K_n}{|\mathcal{F}(n')|}$ is fixed 
for
$\forall n \in \mathcal{F}(n')$, and with the same argument for $W_{n}$ in (\ref{NHSVM}), change of variables gives , $W'_{n'}=\sum_{n \in \mathcal{F}(n')}W_{n}$. Then (\ref{NHSVM}) is a minimization  w.r.t $W'_{n'}$, 
and the minimum is when  $W_{n}= \frac{W'_{n'}}{|\mathcal{F}(n')|}$ for $\forall n\in \mathcal{F}(n')$, plugging
this in (\ref{NHSVM}),
\begin{align}
\lambda \sum_{n\in \mathcal{N}}|\mathcal{F}(n')|&\dfrac{\|W'_{n'}\|^2_2}{|\mathcal{F}(n')|^2} +
\sum_i\max_{y\in \mathcal{Y}}\left (\sum_{n\in y} |\mathcal{F}(n')|\cdot
\ \sqrt{\frac{K_n}{|\mathcal{F}(n')|}}\right. \nonumber \\ \cdot
\frac{W'_{n'}}{|\mathcal{F}(n')|}  
 &-\left. \sum_{n\in y_i}|\mathcal{F}(n')|\cdot\sqrt{\frac{K_{n'}}{|\mathcal{F}(n')|}} \frac{W'_{n'}}{|\mathcal{F}(n')|}\right )\cdot x_i\nonumber
\\+\tilde{\triangle}(y,y_i)\label{plug_W}
\end{align}
By substituting $W_n''=\frac{1}{\sqrt{|\mathcal{F}(n')|}}W'_{n'}$,
\begin{align*}
&\eqref{plug_W}=\lambda \sum_{n} \|W''_n\|^2_2 \\&+\sum_i\max_{y\in\mathcal{Y}}
\left (\sum_{n\in y}  
\sqrt{K_{n'}} W''_n-
\sum_{n\in y_i}\sqrt{K_{n'}} W''_n\right )
\cdot x_i\\&+\tilde{\triangle}(y,y_i) 
\end{align*}

(\ref{NHSVM}),(\ref{beta_1}) for $\mathcal{G}$ are equivalent
to those of $M(\mathcal{G})$, thus two solutions are equivalent with a change of
variables and the decision boundaries are the same.

\qed  
\end{repthm}

\section{Behavior of Shared Frobenius Norm}\label{apx:shared}

We first show a lower bound for $\|\cdot\|_{s}$, $\|\cdot\|_{s,\mathcal{G}}$ which will be useful for the later proofs.
\begin{lem} \label{lower_bound}For $U\in \mathbb{R}^{Y \times d}$,
\begin{align*}
\|U\|_{s,\mathcal{G}}\ge \|U\|_{s}\ge\max_y\|U_y\|_2^{2}
\end{align*}
where $U_y$ is $y$-th row vector of $U$.
\proof
 Let $A\in \mathbb{R}^{Y\times M}, V\in \mathbb{R}^{M \times D}$ be the matrices which attain minimum in 
 $\|U\|_s=\min_{AV=U,\|A\|_{2\rightarrow\infty}\le 1}\|V\|_F$. Since  $A_{r,\cdot} V_{\cdot, c}=U_{r,c}$ and from the cauchy-schwarz,  $\|U_{r,c}\|\le \|A_{r,\cdot}\|_2\cdot \|V_{\cdot, c}\|_2= \|V_{\cdot, c}\|_2$, and if we square both sides and sum over $c$, $\|U_{r,\cdot}\|^2_2\le  \|V\|_F^2=\|U\|^{2}_s$ which holds for all $r$. \qed
\end{lem}
 Following are the detailed descriptions for table \ref{compare_norms} and  the sketch of the proofs. 

\begin{description}
\item 
[Full sharing] 
If  all weights are  same for all classes, i.e., $U=[u\; u\;
\dots\; u]^T\in \mathbb{R}^{Y\times d}$ for $u\in \mathbb{R}^{d}$, and there exists a node $n$ that it
is shared among all $y$, i.e., $\exists n,\forall l,n\in \bar{\mathcal{A}}(l)$, then $\|U\|_{s,\mathcal{G}}^{2}=\|u\|^{2}_2$
whereas $\|U\|^2_F=\|U\|_*^2=Y\cdot\|u\|^{2}_2$.

$\|U\|_{s,\mathcal{G}}^{2}=\|u\|^{2}_2$ can be shown with  matrix  $A=[{\bf{1}_{Y,1}} \;  {\bf{0}_{Y,M-1}}]\in \mathbb{R}^{Y\times M}$  and $V= \begin{bmatrix} u\\ {\bf{0}_{M-1,d}}\end{bmatrix}$  where  ${\bf{n}_{r,c}}\in\mathbb{R}^{r\times c}$ is a matrix with all elements set to $n$. $U=AV$ and the factorization attains the minimum of \eqref{eq:snorm1} since it attains the lower bound from lemma \ref{lower_bound}. $\|U\|_*^2=Y\cdot\|u\|^{2}_2$ is easily shown from the fact that $U$ is a rank one matrix with a singular value of $\sqrt{Y}\cdot\|u\|_2$.

\item[No sharing]\label{lem:no_sharing}  If there is no shared node, i.e., $\forall l,l'\in [Y],l\neq l', \bar{\mathcal{A}}(l)\cap \bar{\mathcal{A}}(l')=\emptyset$, then  $\|U\|_{s,\mathcal{G}}^{2}=\|U\|^{2}_F$.

 To show this, let $A$ and $V$ be the matrices which attain the minimum of (\ref{rho2}).
  $m$-th element of $A_{y}$ is zero  for all $y$ except one and $V_{m,d}$ is nonzero only for one $y$ such
  that $m\in y$.
Therefore, (\ref{rho2})
 decomposes w.r.t $A_{y}$ and $V_{y,d}$, where $V_{y,d}$
 is the $d$-th column vector of
 $V$  taking only for row $y$.
  \begin{align*}
 \min_{AV=U}\|V\|^{2}_F= \sum_y\sum_d\min_{A_yV_{y,d}=U_{y,d}}\|V_{y,d}\|^{2}_2
 \end{align*}
  Given $\|A_{y}\|_2=1$, 
\begin{align*}
\|V_{y,d}\|_2=\|A_{y}\|_2\|V_{y,d}\|_2\ge |A_{y}\cdot V_{y,d}|=|U_{y,d}|\\
\end{align*}
And  let $A_y=V_{y,d}/\|V_{y,d}\|_2$ which attains the lower bound.
\begin{align*}
\therefore\min_{AV=U}\|V\|^{2}_F= \sum_y\sum_d |U_{y,d}|^2=\|U\|^2_F
\end{align*}

\item[Disjoint feature] If  $U=[u_{1}\; u_{2}\;
\dots \; u_Y]^T\in \mathbb{R}^{Y\times d}$ for $l\in [Y]$, $u_{l}\in \mathbb{R}^{d}$, and the support of $w_y$ are all disjoint, i.e., $\forall y_1\neq y_2, \mbox{Supp}(u_{y_1})\cap \mbox{Supp}(u_{y_2})=\emptyset$, then $\|U\|_{s,\mathcal{G}}^{2}=\|U\|_F^{2}=\sum_y\|u_{y}\|^{2}_2$
and $\|U\|_*^2=(\sum_y \|u_y\|_{2})^2$.
 
For $\|\cdot\|_s$, it is similar to  no sharing. The factorization decomposes w.r.t. each column $u$. For the trace norm, since the singular values are invariant to permutations of rows and columns, $U$ can be transformed to a block diagonal matrix by  permutations of rows and columns, and the singular values decompose w.r.t block matrices with corresponding singular values of $\|u_y\|$.

\item[Factor scaling] 
If  $U=[a_1 u\; a_2 u
\dots \; a_Y u]\in \mathbb{R}^{Y\times d}$ for $l\in [L]$, $u\in \mathbb{R}^{d}$, 
 then $\|U\|_{s}^{2}=\max_{l}a_l^2\|u\|_2^{2}$
and $\|U\|_F^2=\|U\|_*^2=\|a\|_2^{2}\cdot\|u\|_2^2. $
 
 Proof is similar to full sharing. For $\|\cdot\|_s$, $A=\dfrac{1}{\max_i a_i}\allowbreak[{[a_1 \; a_2 \; \dots a_Y]^T \; \bf{0}_{Y,M-1}} ]$  and $V=\max_i a_i \begin{bmatrix} u\\ {\bf{0}_{M-1,d}}\end{bmatrix}$  is a feasible solution which attains the minimum in lemma \ref{lower_bound}. For the trace norm, singular values can be easily computed with knowing $U$ is a rank $1$ matrix. 
\section{Convexity of Shared Frobenius Norm optimization}\label{apx:convex}
\begin{replem}{lem:convex}
Optimization \eqref{SSVMp} 
is a convex optimization jointly in $U$ and $\alpha$. 
 \proof
Let $f(U,\alpha)=\sum_n \sum_d f_{n,d}(U_n,\alpha_n)$ where $f_{n,d}={U}_{n,d}^2/\alpha_n$. The Hessian of each $f_{n,d }$ can be calculated easily by differentiating twice. Then, the Hessian is a positive-semidefinite matrix for $\forall \alpha_n\ge0$, since   if $\alpha_n>0$, $\nabla^{2} f_{n,d}=\begin{bmatrix} \frac{\partial^2 f_{n,d}}{(\partial U_{n,d})^2} & \frac{\partial^2 f_{n,d}}{\partial\alpha_n\partial U_{n,d}} \\ \frac{\partial^2 f_{n,d}}{\partial\alpha_n\partial U_{n,d}} & \frac{\partial^2 f_{n,d}}{(\partial\alpha_n )^2}\end{bmatrix}=\frac{2}{\alpha_n}\begin{bmatrix} 1 & -\frac{U_{n,d}}{\alpha_n} \\ -\frac{U_{n,d}}{\alpha_n} & \frac{U_{n,d}^2}{\alpha_n^2}\end{bmatrix}$, and if $\alpha_n=0$ we can assume $\|U_n\|_2=0$ by restricting the domain and  the hessian to be  a zero matrix. Thus, $ \underset{n}{\sum} \dfrac{\|U_n\|^2_2}{\alpha_n}$ is a convex function jointly in $U_n$ and $\alpha_n$, and the lemma follows from the fact that the rest of the objective function in \eqref{SSVMp} is convex in $U_n$. \qed
\end{replem} 
\section{Closed form optimization of $\alpha$}
\label{apx:tree_shared}
\begin{replem}{lem:tree_shared} For a tree structure $\mathcal{G}$, algorithm \ref{argmin_alpha} finds optimal $\alpha$ in \eqref{optimize_alpha} in $O(M)$ in a closed form. 
 \proof
Let $f(n,l)=\min_{\sum_{n\in D(n)}\alpha_n\le l} \allowbreak\sum_{n\in \bar{D}(n)}\dfrac{\|U_n\|^2_2}{\alpha_n}$
where $\bar{D}$ denotes the union set of $\{n\}$ and descendent nodes of $n$.
the following recursive relationship
holds, since  $\mathcal{G}$ has a tree structure. 
\begin{align} 
f(n,l)=\begin{cases} \dfrac{\|U_n\|^2_2}{l} & \mbox{if $n$ is a leaf node} \\ \min_{0<k<1}\dfrac{\|U_n\|^2_2}{l\cdot k}\\+\sum_{c\in C(n)}f(c,l(1-k))  &  otherwise\end{cases}
\label{recursive}
\end{align}
If $n$ is a parent node of  leaf nodes,
\begin{align}
f(n,l)= \min_{0<k<l}\dfrac{B_1}{l\cdot k}+\dfrac{B_2}{l(1-k)}
\label{eq:min_k}
\end{align}
where $C(n)$ denotes the set of children nodes of $n$, $B_1=\|U_n\|^2_2$ and
$B_2=\sum_{c\in C(n)}\|U_{c}\|^2_2$. 
This has a
closed form solution, 
\begin{align}
f(n,l)= 
\frac{1}{l}(\sqrt{B_{1}}+\sqrt{B_{2}})^2 
\label{leaf_f}
\end{align}
and the minimum is attained at $k=\frac{\sqrt{B_1}}{\sqrt{B_1}+\sqrt{B_2}}$.
For nodes $p$ of $n$, $f(p,l)$
will also have a form of (\ref{leaf_f}), since  the equation (\ref{leaf_f})  has a form of leaf node,  and the recursive relationship
(\ref{recursive}) holds. We  continue this process until the root node $r$ is reached,
and $f(r,1)$ is the optimum. The optimal $\alpha$ can be calculate backward. 

\qed
\end{replem}

%

\end{description}

\end{document}